\documentclass{article}

 \usepackage[preprint]{neurips_2020}

\usepackage{times}
\usepackage{latexsym}
\usepackage{booktabs}
\usepackage{graphicx}
\usepackage{multirow}
\usepackage{makecell}
\usepackage{amsfonts}
\usepackage{amsmath}
\usepackage{comment}
\usepackage{xcolor}
\usepackage{todonotes}

\title{Entailment as Few-Shot Learner}

\author{%
  Sinong Wang, Han Fang, Madian Khabsa, Hanzi Mao, Hao Ma\\\\
  Facebook AI, USA\\
 \texttt{sinongwang@fb.com}\\
}

\begin{document}

\maketitle

\begin{abstract}
Large pre-trained language models (LMs) have demonstrated remarkable ability as few-shot learners. However, their success hinges largely on scaling model parameters to a degree that makes it challenging to train and serve. 
In this paper, we propose a new approach, named as EFL, that can turn small LMs into better few-shot learners. The key idea of this approach is to reformulate potential NLP task into an entailment one, and then fine-tune the model with as little as 8 examples. We further demonstrate our proposed method can be: (i) naturally combined with an unsupervised contrastive learning-based data augmentation method; (ii) easily extended to multilingual few-shot learning. A systematic evaluation on 18 standard NLP tasks demonstrates that this approach improves the various existing SOTA few-shot learning methods by 12\%, and yields competitive few-shot performance with 500 times larger models, such as GPT-3.

\end{abstract}

\section{Introduction}

Recent improvements in NLP models relied largely on the pre-training and fine-tuning paradigm, wherein a language model is first pre-trained on a massive text corpora, followed by finetuning on the downstream tasks~\citep{devlin2019bert,liu2019roberta,raffel2020exploring,lewis2020bart}. The performance of the model varies depending on the tasks, and the number of available training examples. However, in practice, there is an overly large number of domains, tasks, and languages, and scaling to a new problem space will require additional labeled data. This leads to an important research area, \textit{few-shot learning}, which assumes accessing to only a small number of labeled examples.

Different from this pre-training/fine-tuning paradigm, GPT-3~\citep{brown2020language} showed that pre-training alone combined with a sufficiently large parameter size yields a model that is capable of achieving nearly SOTA results when prompted with the training examples. More importantly, GPT-3 demonstrated that the number of examples needed for promoting can be as small as one, and performance can be significantly improved when using 16 examples (per class) in a few-shot manner. However, the performance of the model depended largely on the number of parameters, and scaling to 175 billion parameters was imperative to achieve these results. Unfortunately, this scale introduces significant challenges for both training and serving.


Various methods have been proposed to equip smaller language models with few-shot capabilities. One alternative approach is to follow the Masked Language Modeling (MLM) paradigm~\citep{devlin2019bert}, and reformulate downstream tasks as similar cloze questions (e.g., by appending phrases such as ``the correct answer is "), allowing pre-trained LMs to predict the correct label by reusing MLM head~\citep{schick2020exploiting,schick2020s}. The performance of these techniques are limited when the downstream tasks have different data distribution from the pre-trained text corpora. For example, the work~\citep{gao2020making} shows that pre-trained LM performs worse in some tasks such as linguistic acceptability task and natural language inference task since pre-trained LM barely saw this data distribution.

\begin{figure*}[t]
  \centering
  \includegraphics[width=\linewidth]{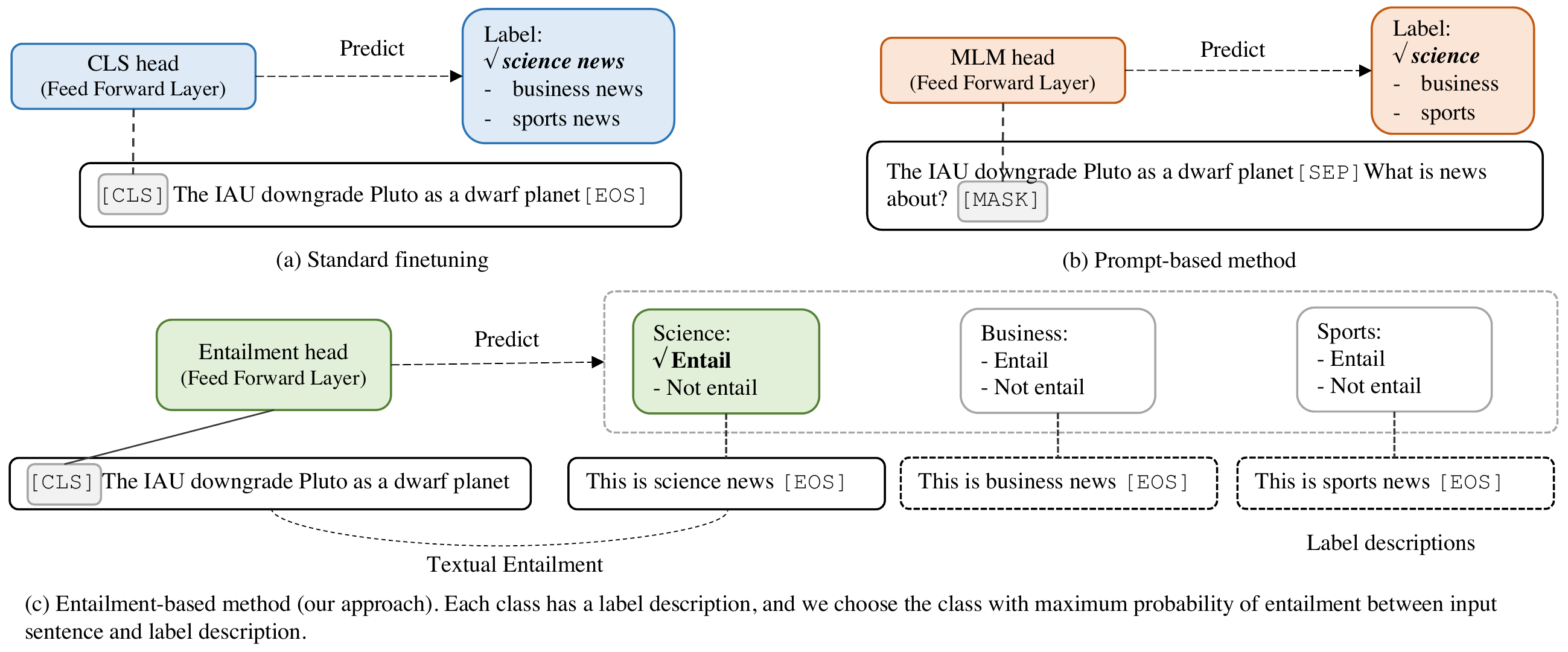}
  \caption{An illustration of (a) standard fine-tuning of a classification problem; (b) prompt-based method; (c) our proposed method using entailment-based fine-tuning. Compared to prompt-based methods, the key difference of this approach is reformulating tasks as entailment task instead of cloze questions and design fine-grained label descriptions instead of a single task description.}
  \label{fig:architecture}
\end{figure*}

In this paper, we propose a new approach, in which NLP tasks are first reformulated as a textual entailment task. The assumption here is that entailment can be used as unified method to model all classification tasks\footnote{This approach does not extend to generative tasks, such as translation or summarization}. As illustrated in Figure~\ref{fig:architecture}(c), the key idea is to convert the class label into a natural language sentence which can be used to describe the label, and determine if the example entails the label description. For example, we can reformulate a sentiment classification input/label pair: [$x:$ \textit{I am in love with these actors}\texttt{[EOS]}, $y:$ \textit{positive}] as following textual entailment sample: [$x:$ \textit{I am in love with these actors}\texttt{[SEP]}\textit{This is a great movie}\texttt{[EOS]}, $y:$ \textit{entailment}]. 

By converting tasks into this entailment style, we demonstrate that standard pre-trained language models are very effective few-shot learners. Another benefit is that, since various tasks are reformulated as a sentence-pair entailment task, we can utilize contrastive learning to construct pairwise augmented data to further improve the few-shot performance. Experiments on wide variety of tasks including 8 tasks from GLUE benchmark, SNLI, BoolQ from superGLUE, and 8 other popular sentence classification tasks show that such off-the-shelf entailment models can improve few-shot performance by 12\% compared to various few-shot learning methods. Our methods also established 1.9pt absolute improvement in full training dataset compared to standard fine-tuning of RoBERTa model. We further extend this method to multilingual few-shot learning, which also shows average 19pt improvement compared to standard fine-tuning method.

The model's striking ability to perform few-shot learning could be due to entailment being a true language understanding task, that once the model is capable of performing it correctly, it can easily apply this knowledge to other tasks which are framed as such. While the proposed approach still requires fine-tuning the model for each task, it does not require a prohibitively large language model to achieve strong performance. Moreover, entailment models are widely accessible to everyone to download and fine-tune through various repositories, thus it democratizes few-shot learners and does not limit it to commercial black-box APIs. 


\section{Related Work}

There have been many research studies on improving the few-shot learning performance of a pre-trained language model:

\textbf{Language modeling with demonstrations:} The series of GPT works~\citep{radford2019language,brown2020language} proposed to add a task description (prompt) and annotated examples as demonstration to enable few-shot learning, which has been commonly applied to classification~\citep{puri2019zero}, QA, commonsense knowledge mining~\citep{davison2019commonsense}. It is also explored for probing the knowledge contained within pre-trained LMs.

\textbf{Task reformulation:} Language model is usually pre-trained with a Masked Language Modeling (MLM) objective, which is motivated by Cloze task in~\citep{taylor1953cloze}. There have been several works reformulating few-shot learning tasks as cloze questions to reuse pre-trained LM such as LM prompt~\citep{jiang2020can}, PET~\citep{radford2019language,schick2020exploiting}, and recent LM-BFF~\citep{gao2020making}. It shows a pre-trained LM can achieve non-trivial performance with few annotated samples. There are also some other works transforming NLP tasks as generative QA tasks~\citep{puri2019zero}.

\textbf{Intermediate training:} The work by~\citep{phang2018sentence} shows that supplementing pre-trained LMs with further training on data-rich supervised tasks can obtain additional performance improvements on the GLUE benchmark. More recently, this approach has been further improved by a matching-based few-shot learning method~\citep{yin2020universal}.

\textbf{General techniques:} There are several general techniques to improve the few-shot learning performance, including: (i) optimization and regularization techniques during the fine-tuning~\citep{howard2018universal,lee2019mixout,zhang2020revisiting}, (ii) semi-supervised learning to augment training data~\citep{xie2020unsupervised}, and (iii) supervised contrastive learning as additional objective~\citep{gunel2020supervised}. We anticipate that these studies are largely complementary to ours.

Comparing to existing prompt-based few-shot learning methods~\citep{brown2020language,schick2020exploiting,gao2020making}, the key differences of our proposed method are: (i) our approach reformulates NLP tasks as textual entailment instead of cloze questions; (ii) provide label-specific descriptions for each class instead of single task description.

\section{Framework of Entailment Training}

In this section, we introduce our motivation and framework of entailment training.

\begin{figure*}[t]
  \centering
  \includegraphics[width=\linewidth]{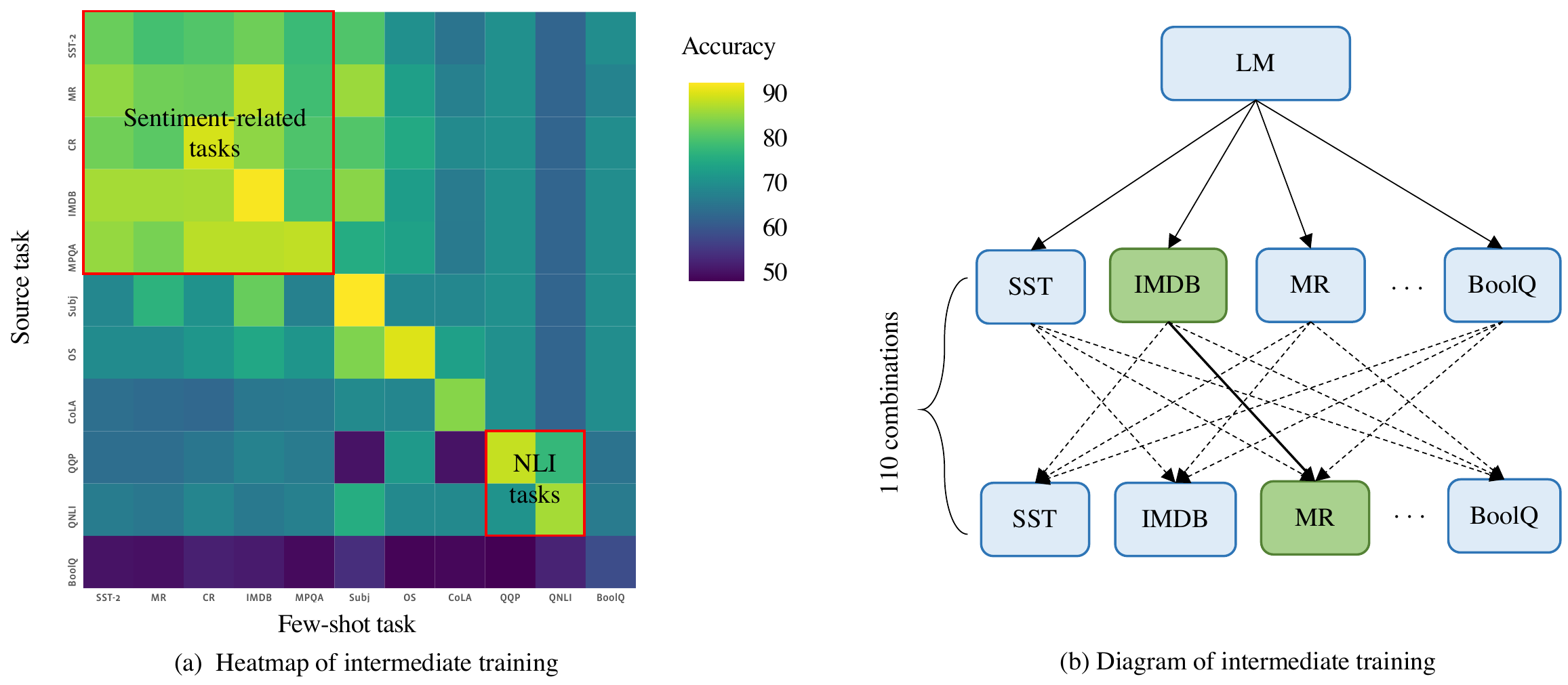}
  \caption{An illustration of intermediate training: first fine-tune the pre-trained LM on a source task with rich annotated data, then further fine-tune the model with 8 annotated samples per class in target task. Figure (a) illustrates the accuracy heatmap between 11 tasks. It shows that sentiment-related task and NLI task have better transferability among each other and non-trivial few-shot results, while other task type such as topic classification, subjectivity task has marginal improvements. Figure (b) illustrates the diagram of this approach, which requires 110 times fine-tuning runs to fine the best results among 11 tasks.}
  \label{fig:motivation}
\end{figure*}

\subsection{Motivation}

We assume there exists a pre-trained language model $\mathcal{M}$ and a new downstream task $\mathcal{T}$ with label space $\mathcal{Y}$. Suppose that the training dataset $\mathcal{D}_{\text{train}} = \{D_1, D_2, \ldots, D_{|\mathcal{Y}|}\}$, where $D_k=\{x_i, y_i\}_{i=1}^{K}, k\in \mathcal{Y}$ and $K$ is the number of training data for each class. In this paper, we set $K=8$ as default. Note that label space $\mathcal{Y}$ can be both discrete and continuous space, which corresponds to classification task and regression task, separately. The objective is to develop a task-agnostic training methodology of language model $\mathcal{M}$ on limited training data $\mathcal{D}_{\text{train}}$ such that it generalizes well to examples on test set $\mathcal{D}_{\text{test}}$. 

In the standard NLP tasks, the input $x_{\text{in}}$ is usually a single sentence $S_1$ (classification, regression) or a sentence pair $(S_1, S_2)$ (natural language inference, QA). In the standard fine-tuning approach, given a pre-trained language model $\mathcal{M}$, we usually take
$x_{\text{in}}=\texttt{[CLS]}S_1\texttt{[EOS]}$ or $x_{\text{in}}=\texttt{[CLS]}S_1\texttt{[SEP]}S_2\texttt{[EOS]}$, and map $x_{\text{in}}$ into a sequence of hidden vectors $\mathcal{M}(x_{\text{in}})$. Then we can predict the output classes via a downstream classification head  $\texttt{softmax}(\mathbf{W}\mathbf{h}_{\texttt{[CLS]}})$, where $\mathbf{h}_{\texttt{[CLS]}}$ is the hidden vector of \texttt{[CLS]} token. During the fine-tuning phase, the parameters of both language model $\mathcal{M}$ and classification head $\mathbf{W}$ will be updated to minimize the cross entropy between \texttt{softmax} output and correct label~\footnote{For regression task, we will minimize the mean square error.}.

The key challenge of few-shot learning is that we have limited training data, while a huge amount of model parameters need to be updated, e.g., BERT-large has 340M parameters. To resolve this challenge, one alternative approach is to reuse the language model parameters if the model is already pre-trained with a similar task. For example, in Figure~\ref{fig:motivation}, we conduct a two-stage experiment across 11 NLP tasks (details of data stats can be seen in Appendix~\ref{table:data_statistics}). We first fine-tune a RoBERTa-Large~\citep{liu2019roberta} model on the full dataset of one task, then we load the model parameters $\mathcal{M}$ and continue fine-tune on another task in a few-shot setting (8 training data per label class in each task, and fine-tuned with 5 random split). We report the average accuracy on full $\mathcal{D}_{\text{test}}$. As illustrated in Figure~\ref{fig:motivation}(a), we can achieve reasonable few-shot learning performance, i.e. 85\% accuracy, on most sentiment tasks such as fine-tuning on IMDB first, then fine-tune on MR. However, in other tasks such as OS, CoLA and NLI tasks, all the source tasks brings marginal improvement.

Moreover, the above approach is not task-agnostic. It requires substantial amount of sweeping over available source tasks (Figure~\ref{fig:motivation}(b)), and the domain-knowledge of the downstream task. One natural question is that \textit{can we build a single language model $\mathcal{M}$ that is task-agnostic and generalize well in few-shot setting?}

\subsection{Framework}

To answer this question, we propose a new framework in which  NLP tasks are transformed as a textual entailment task. We refer to this framework as \textbf{EFL}, short for \textbf{E}ntailment as \textbf{F}ew-shot \textbf{L}earner. For instance, we can reformulate a sentiment classification task as a textual entailment one with an input sentence $S_1$ as
\begin{align}
&x_{in}=\texttt{[CLS]}S_1\texttt{[SEP]}S_2\texttt{[EOS]}, \text{ where } S_2=\text{This indicates positive user sentiment,}\label{eq:enta_eq}
\end{align}
and let the language model $\mathcal{M}$ to determine the if input sentence $S_1$ entails the label description $S_2$. It is important to note that this approach consolidates all the one-sentence classification/regression tasks and sentence-pair tasks into a unified template $\texttt{softmax}[\mathbf{W}\mathcal{M}(x_{\text{in}})_{\texttt{[CLS]}}]$, where $x_{\text{in}}$ is defined in (\ref{eq:enta_eq}). Then we can reduce the gap between traditional MLM pre-training and fine-tuning by further \textit{pre-training} the language model with existing entailment tasks or reformulating other tasks as entailment task such that we can reuse the model parameters $\mathbf{W}$ and $\mathcal{M}$.

\begin{table*}[t]
\centering
\resizebox{0.97\textwidth}{!}{
\centering
\begin{tabular}{lc|lc}
\toprule
Label Template  & Accuracy & Label Template  & Accuracy \\\midrule
\textit{SST-2 sentiment classification} & & \textit{OS hate speech classification} \\\midrule
This review indicates positive experience & 89.8 (1.2) & This tweet contains hate speech & 82.6 (2.1)\\
This review indicates great experience & 90.6 (0.7) & This tweet contains offensive words & \textbf{83.2} (2.6)\\
It is great & 90.8 (1.0) & It is hate speech & 76.6 (3.5)\\
It is great movie & \textbf{91.0} (0.6) & It is offensive speech & 79.3  (2.4)\\
It is cute &  87.4 (2.1) & It is peace and love & 74.3 (4.7) \\
\bottomrule
\end{tabular}}
\caption{The impact of label descriptions on our proposed EFL method with $K=8$. The performance is relative stable if the designed label descriptions are close to the problem definition, but degenerates if choosing semantic-unrelated label descriptions.}
\label{table:label_ablation}
\end{table*}

In the binary classification/regression task, we simply choose a label description $p$ as second sentence $S_2$, and fine-tune the model on $\mathcal{D}_{\text{train}}=\{(x_i, p, y_i)\}_{i=1}^{K|\mathcal{Y}|}$. The remaining challenge is to choose the label descriptions for different tasks. Similar as the existing work ~\citep{schick2020s}, we hand-craft the label descriptions for each task. Details can be seen in Table~\ref{table:label_prompt}. The general design principle is to intuitively choose some simple words to describe the labels. For instance, in the sentiment classification task, we can choose \emph{This movie is great} or \emph{This review indicates positive user experience} as label description. In Table~\ref{table:label_ablation}, we show that our method actually requires minimum domain knowledge to choose the label descriptions. Note that in sentence-pair task such as NLI task, this approach is degenerating to the standard fine-tuning method.

In the multi-class classification task, we choose a label description $p_k$ for each class $k\in\mathcal{Y}$. Suppose that the dataset $\mathcal{D}_{\text{train}}=\{\mathcal{D}_i\}_{i=1}^{|\mathcal{Y}|}$, where $D_i=\{(x_j,y_j)\}_{j=1}^{k}$ is the data for $i$th class. Based on the designed label descriptions $\{p_k\}_{k\in\mathcal{Y}}$, we reshape each dataset as an entailment dataset $\hat{D}_k=\{(x_j,p_k,y_j)\}_{j=1}^{K}$. To accommodate the entailment training, we reshape the training dataset as 
\begin{equation}
\mathcal{\hat{D}}_{\text{train}}=\{(\hat{D}_i,\hat{D}_{-i})\}_{i=1}^{|\mathcal{Y}|},
\end{equation}
where $\hat{D}_{-i}$ is a random subset of $\{\hat{D}_j\}_{j\neq i, j\in\mathcal{Y}}$ with 
$|\hat{D}_{-i}|=K$. 
For instance, in AG news classification~\citep{zhang2015character}, we can formulate a simple set of label descriptions \{\textit{This is worlds news}, \textit{This is business news}, \textit{This is sports news}, \textit{This is science news}\} for total 4 classes. Note that this approach requires $|\mathcal{Y}|$ forward passes during inference time, while standard fine-tuning only requires one forward pass to get the prediction of all the classes. Actually, the inference speed can be easily improved if we reuse the siamese transformer networks~\citep{reimers2019sentence}.
\begin{table*}[t]
\resizebox{\textwidth}{!}{
\begin{tabular}{cccccc}
\toprule
\textbf{Dataset} & $|L|$ & $\#$Train & $\#$Test & \textbf{Type} & \textbf{Labels Descriptions}\\
\midrule
\textit{One-sentence Task}\\\midrule
SST-2 & 2 & 67,349 & 872 & sentiment & positive, negative experience\\
MR & 2 & 8,662 & 2,000 & sentiment & positive, negative experience \\
CR & 2 & 1,775 & 2,000 & sentiment & positive, negative experience \\
MPQA & 2 & 8,606 & 2,000 & opinion polarity & positive/negative opinion \\
Subj & 2 & 8,000 & 2,000 & subjectivity &  subjective/objective\\
OS & 2 & 3,000 & 3,000 & topic classification & hate speech/benign twitters\\
IMDB & 2 & 25000 & 25000 & sentiment & positive/negative user experience \\
CoLA & 2 & 8,551 & 1,042 & acceptability &  grammatical, not grammatical \\
TREC & 6 & 5,452 & 500 & question classification & abbr., entity, description, human, loc., num.\\
Yelp & 5 & 600k & 1,000 & sentiment & 5-scale user sentiments \\
AG News & 4 & 4,000 & 4,000 & topic classification & world, sports, business, science news \\\midrule
\textit{Sentence-pair task}\\\midrule
QQP & 2 & 363,846 & 40,431 & paraphrase & equivalent, not\_equivalent \\
MRPC & 2 & 3,668 & 408 & paraphrase & equivalent, not\_equivalent  \\
QNLI & 2 & 104,743 & 5,463 & NLI &  entailment, not\_entailment \\
SNLI & 3 & 549,367 & 9,842 & NLI &  entailment, neutral, contradiction \\
RTE & 2 & 2,490 & 277 & NLI & entailment, not\_entailment \\
STS-B & 2 & 5,749 & 1,500 & sentence similarity & $[0, 5]$ continuous score \\
BoolQ & 2 & 9,427 & 3,270 & QA & Yes/No \\
\bottomrule
\end{tabular}}
\caption{The datasets evaluated in this work. $|L|$: $\#$ of classes for classification tasks. Note that we only sample $D_{\mbox{train}}$ with $K=8$ examples from the original training set and evaluate using full dev dataset in our few-shot experiments }
\label{table:data_statistics}
\end{table*}
\section{Experiment}

In this section, we present experiment results of our proposed approach and various existing few-shot learning methods. We also evaluate the impact of training data size and pre-trained model size on those methods.

\subsection{Data Statistics}

We conduct a comprehensive study across 18 NLP tasks (see
Table~\ref{table:data_statistics}), which covers sentiment analysis, topic classification, natural language inference, paraphrases, sentence similarity and QA. Our evaluation consists of 8 tasks from the GLUE benchmark~\citep{wang2018glue}, SNLI~\citep{bowman2015large}, BoolQ~\citep{clark2019boolq} from SuperGLUE and 8 other popular sentence classification tasks (Yelp, MR, CR, MPQA, Subj, IMDB, AG News, OS, Trec). See Appendix A for details. We divide our tasks into two categories: (i) \emph{One-sentence task:} The input of this task is a single sentence, the goal is to predict the binary or multi-class label. For example, predicting whether user sentiment is positive or negative. 
(ii) \emph{Two-sentence task:} The goal of this task is to make prediction based on a sentence pair. For example, predicting similarity of two input sentences . 

\subsection{Evaluation and Training Protocol}

It is well-known that fine-tuning and evaluating on small datasets can suffer from instability and results may change dramatically given a new split of data. To address this, existing work~\citep{gao2020making} proposed to measure the average performance across 5 different randomly sampled train and dev splits, using a fixed set of seeds. To further reduce the instability of evaluation, we measure the performance on full test dataset instead of randomly downsampled 5 small dev splits. In all of our experiment, we use 8 training samples per class, and the size of evaluation data is listed in TABLE~\ref{table:data_statistics}. We also follows the principle in~\citep{schick2020exploiting,schick2020s} and assume no access to a development set and adopt a fixed set of
hyper-parameters based on practical considerations.

For fair comparison, we use the same label descriptions for both prompt-based fine-tuning and entailment-based fine-tuning. For each experiment, we run 5 experiments with 5 different training split and report the average results and standard deviations. We experiment all the methods using a RoBERTa-large model~\citep{liu2019roberta}, which is a 24-layer, 16-heads transformer model with 355M model parameters. Details of label descriptions and hyper-parameter setup can be seen in Appendix~\ref{appendix:experiment}.

\begin{table*}[t]
\renewcommand{\arraystretch}{1.1}
\resizebox{\textwidth}{!}{
\begin{tabular}{lcccccccc}
\toprule
 \textbf{Method} & \textbf{SST-2} & \textbf{MR} & \textbf{CR} & \textbf{MPQA} & \textbf{Subj} & \textbf{IMDB} & \textbf{OS}&  \textbf{CoLA}\\ 
  & (Acc) & (Acc) & (Acc) & (Acc) & (Acc) & (Acc) & (Acc) & (Acc.) \\
\midrule
\multicolumn{2}{l}{\textit{Full training dataset}}\\
\midrule
Majority & 50.9 & 50.0 & 50.0 & 50.0 & 50.0  & 50.0 & 66.8 & 69.1 \\
 Fine-tuning & 96.4 (0.2) & 91.6 (0.2) & 91.8 (0.7) & 89.4 (0.9) & \textbf{97.4} (0.1) & 96.1 (0.2) & 94.3 (0.1) & 86.2 (1.6)  \\
 EFL & \textbf{96.9} (0.2) & \textbf{92.5} (0.1) & \textbf{92.5} (0.4)&  \textbf{90.8} (0.4) & 97.1 (0.2)& 96.1 (0.2) & \textbf{95.1} (0.1) & \textbf{86.4} (0.5)\\
 \midrule
\multicolumn{2}{l}{\textit{Few-shot with $K$=8}}\\
\midrule
Fine-tuning & 60.5 (3.1) & 60.3 (7.5) & 61.9 (5.1) & 59.0 (3.4) & 78.3 (8.2) & 73.5 (7.8) & 70.0 (5.4) & \textbf{70.0} (0.9) \\
Stilts-NLI  & 64.5 (5.4) & 63.3 (4.3) & 68.8 (7.0) & 67.7 (5.7) & 69.1 (5.6) & 59.7 (3.6) & 62.4 (5.7) & 66.3 (3.5)\\
Stilts-Close & 85.5 (2.3) & 81.9 (4.8) & 84.7 (5.4) & 76.6 (1.3) & \textbf{83.2} (3.4) & 86.9 (4.0) & 68.1 (1.4) & 62.2 (3.9) \\
LM-BFF & 79.9 (6.0) & 85.4 (3.9) & 88.6 (2.3) & 74.2 (1.2) & 81.6 (6.2) & \textbf{92.0} (0.5) & 71.0 (3.3) & 69.5 (0.5)\\
EFL wo PT & 58.6 (6.7) & 55.1 (1.1) & 64.0 (5.6) & 60.4 (3.7) & 77.9 (6.1) & 75.5 (9.6) & 71.2 (3.2) & 70.4 (1.9) \\
EFL & \textbf{90.8} (1.0) & \textbf{86.2} (0.8) & \textbf{92.3} (0.4) & \textbf{87.0} (0.6) & 80.0 (5.4) & 87.1 (1.2) & \textbf{76.6} (3.5) & 69.4 (0.9)\\
\bottomrule
\end{tabular}}
\caption{Our main few-shot learning results using RoBERTa-large on one sentence task. All the results are evaluated on full test sets and averaged across 5 different training sets.}
\label{table:few_shot_16_1}
\end{table*}
\subsection{Main Results}

We conduct both experiments on full-shot and few-shot scenarios, and compare our proposed methods with various baselines, including: 

\textbf{Majority}: simply take the most frequent class (measured on the full test dataset).

\textbf{Standard FT}: standard fine-tuning of pre-trained language model on full/few-shot dataset with cross entropy loss (or mean square error for regression task).

\textbf{STILTS}~\citep{phang2018sentence}: STILTS is widely used
technique to improve the performance of a target problem by creating intermediate training task. Traditionally, people pre-train $\mathcal{M}$ on MNLI and then fine-tune on target data. We also create a new stronger version, named as STILTS-close, which pre-train $\mathcal{M}$ across the tasks in Table~\ref{table:data_statistics}, and report average results of best 4 source tasks.

\textbf{LM-BFF}~\citep{gao2020making}: add prompts in each input sentence (pair) and replace keywords by masked tokens, then predict masked tokens by reusing the output layer of pre-trained mask language model. This method has show superior performance across GPT-3 in-context learning and PET~\citep{schick2020exploiting}.

\textbf{EFL}: Our proposed method, which transforms label descriptions as a input sentence and reformulate original classification/regression task as a entailment task. We also create two version of this methods: (i) EFL wo PT refers to directly fine-tune pre-trained language model; (ii) EFL refers to first train on MNLI task, then fine-tune on downstream tasks.

\begin{table*}[t]
\centering
\renewcommand{\arraystretch}{1.1}
\resizebox{0.95\textwidth}{!}{
\begin{tabular}{lccccccc}
\toprule
 \textbf{Method}& \textbf{QQP} & \textbf{QNLI} & \textbf{SNLI} & \textbf{RTE} & \textbf{MRPC} & \textbf{STS-B} & \textbf{BoolQ}\\ 
 & (F1) & (Acc) & (Acc) & (Acc) & (F1) & (Pear.) & (Acc) \\
\midrule
\multicolumn{2}{l}{\textit{Full training dataset}}\\\midrule
Majority & 0.0  & 50.5 & 33.8 & 52.7 & 81.2 & -1  & 62.3 \\
Fine-tuning & 89.0 (0.1) & 92.9 (0.2) & 77.6 (8.8) & 90.2 (0.6) & 89.9 (1.7) & 86.1 (0.5) & \textbf{86.1} (0.5)\\
EFL & \textbf{89.2} (0.1)  & \textbf{94.5} (0.1) & \textbf{93.1} (0.2) & \textbf{90.5} (0.4) & \textbf{91.0} (0.8) & 91.8 (0.3) & 86.0 (0.2) \\\midrule
\multicolumn{2}{l}{\textit{Few-shot with $K$=8}}\\\midrule
Fine-tuning & 58.8 (9.9) & 52.7 (1.8) & 38.4 (1.3) & 55.0 (1.3) & 76.1 (3.9) & 24.5 (8.4) & 60.8 (2.8)\\
LM-BFF & \textbf{68.2} (1.2)  & 61.8 (3.2) & 52.0 (1.7) & 63.3 (2.1) & \textbf{78.5} (2.3) & 66.0 (3.2)  & 71.2 (3.5) \\
EFL & 67.3 (2.6)  & \textbf{68.0} (3.4) & \textbf{81.0} (1.1) & \textbf{85.8} (0.9) & 76.2 (1.3) & \textbf{71.0} (1.3)  & \textbf{73.9} (1.8)\\
\bottomrule
\end{tabular}}
\caption{Our main few-shot learning results using RoBERTa-large on NLI, paraphrase, similarity and QA tasks. All the results are evaluated on full dev sets and averaged across 5 different training sets.}
\label{table:few_shot_16_2}
\end{table*}

Table~\ref{table:few_shot_16_1} and Table~\ref{table:few_shot_16_2} shows the main results of 15 NLP tasks. We observe that our proposed method EFL greatly outperforms standard fine-tuning, LM-BFF, Stilts-NLI and even a very strong baseline Stilts-Close, which assumes the access of development set to sweep the choice of upstream task. In few-shot scenario, $K$=8, EFL achieves average \textbf{8.2\%} (up to 55\%) improvements across 15 tasks among those existing methods. The only exception is the CoLA (the linguistic acceptability task). It is might due to that MLM pre-trainng and entailment training did not see this type of data distribution before. In the full training dataset, EFL shows around \textbf{1.9pt} average improvements compared to standard fine-tuning of RoBERTa-Large\footnote{Note that we have very large improvement on SNLI, if we remove that benchmark, the average improvement is 1.0pt.}. These results identify the effectiveness of the proposed entailment-based method as a better approach for few-shot learning, moreover, a unified approach for various NLP tasks. 

Furthermore, we observe that if we remove the entailment training step of EFL and directly fine-tune a language model as an entailment task, the performance drops significantly in 8-shot scenario, even worse compared to the standard fine-tuning. This is an interesting case as we will show that, in the next section, EFL without entailment training can also perform well if we increase $K$ to 16 or 32. Note that the overall label description we used in our experiments is default choice of prompt-based method~\citep{gao2020making}. As shown in Table~\ref{table:label_ablation}, we observe that our method can achieve even better performance if we optimize the label descriptions.

\begin{table*}[htb]
\centering
\renewcommand{\arraystretch}{1.1}
\resizebox{\textwidth}{!}{
\begin{tabular}{lcccccccc}
\toprule
\textbf{Datasets} & \multicolumn{2}{c}{Majority} &\multicolumn{2}{c}{Standard FT}  & \multicolumn{2}{c}{LM-BFF} & \multicolumn{2}{c}{EFL}\\ 
& Acc& Macro-F1 & Acc & Macro-F1 & Acc & Macro-F1 & Acc & Macro-F1\\
\midrule
AG News & 25.0 & 0.08 & 81.5 (2.9) & 51.1 (9.9) & 85.4 (2.3) & 53.5 (5.3)  & 
\textbf{86.1} (2.9) & \textbf{79.5} (3.8) \\
Yelp & 41.1 & 11.7 & 49.8 (4.3) & 40.1 (3.0) & 64.5 (2.5) & 40.5 (2.9) & \textbf{64.9} (2.6) & \textbf{42.6} (3.9)\\
Trec & 27.6 & 8.65 & 53.0 (9.3) & 23.2 (0.2) & \textbf{81.5} (5.7) & 49.1 (1.5) & 80.9 (3.1) & \textbf{69.0} (4.9)\\
\bottomrule
\end{tabular}}
\caption{Our main few-shot learning results using RoBERTa-large on multi-class tasks. Besides accuracy metric, we also report macro-F1 metric since these multi-class datasets are unbalanced across different classes.}
\label{table:multi_class}
\end{table*}
Table~\ref{table:multi_class} shows the main results of three multi-class benchmarks. Both LM-BFF and our proposed method have better performance than standard fine-tuning. Each multi-class benchmark has unbalanced data across different classes, using accuracy metric is unrepresentative in this scenario. An interesting observation is that, when we use Macro-F1 metric (average F1 across each class) to measure the performance, our proposed method has significantly better performance than other methods. It implies that EFL has better performance in some low-resource classes.

\begin{figure*}[t]
  \centering
  \includegraphics[width=0.9\linewidth]{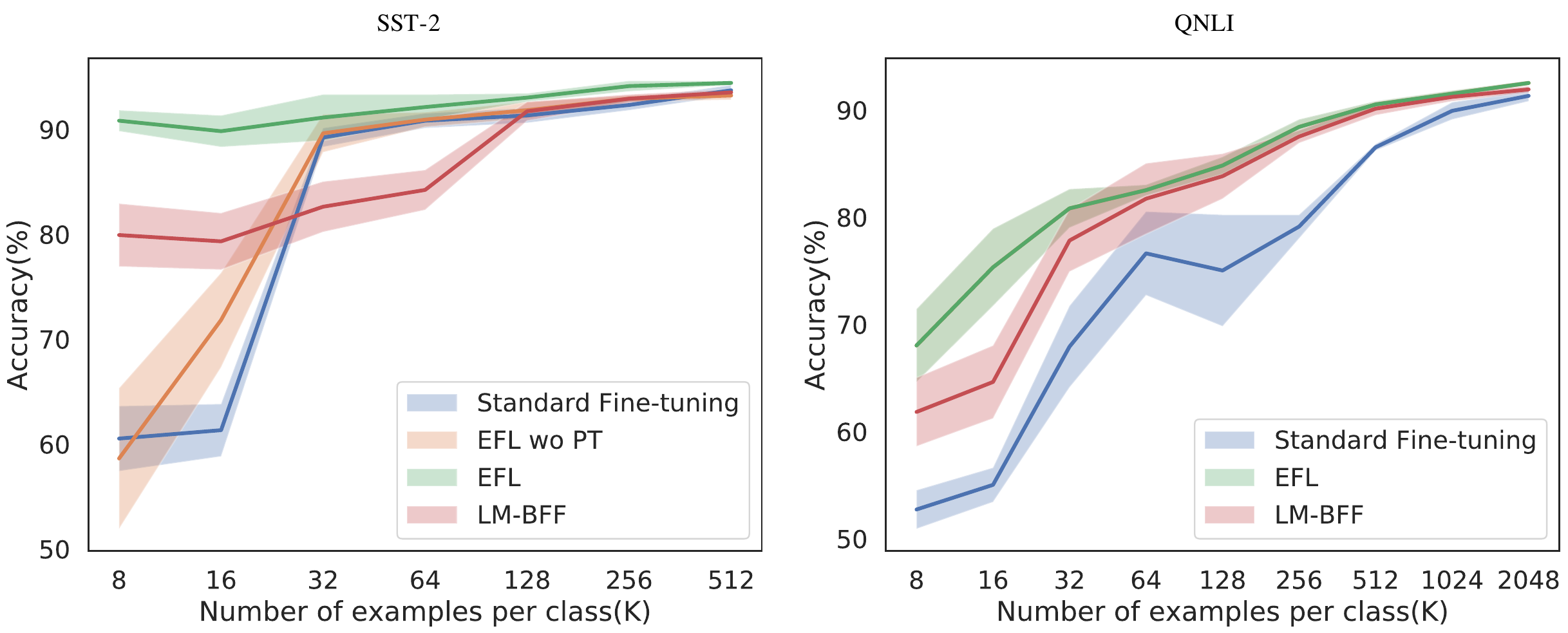}
  \caption{Sample efficiency of standard fine-tuning, LM-BFF, our proposed EFL and EFL without entailment pre-training as a function of $K$ (number of instances per class).}
  \label{fig:sample_efficiency}
\end{figure*}
\begin{figure*}[ht]
  \centering
  \includegraphics[width=\linewidth]{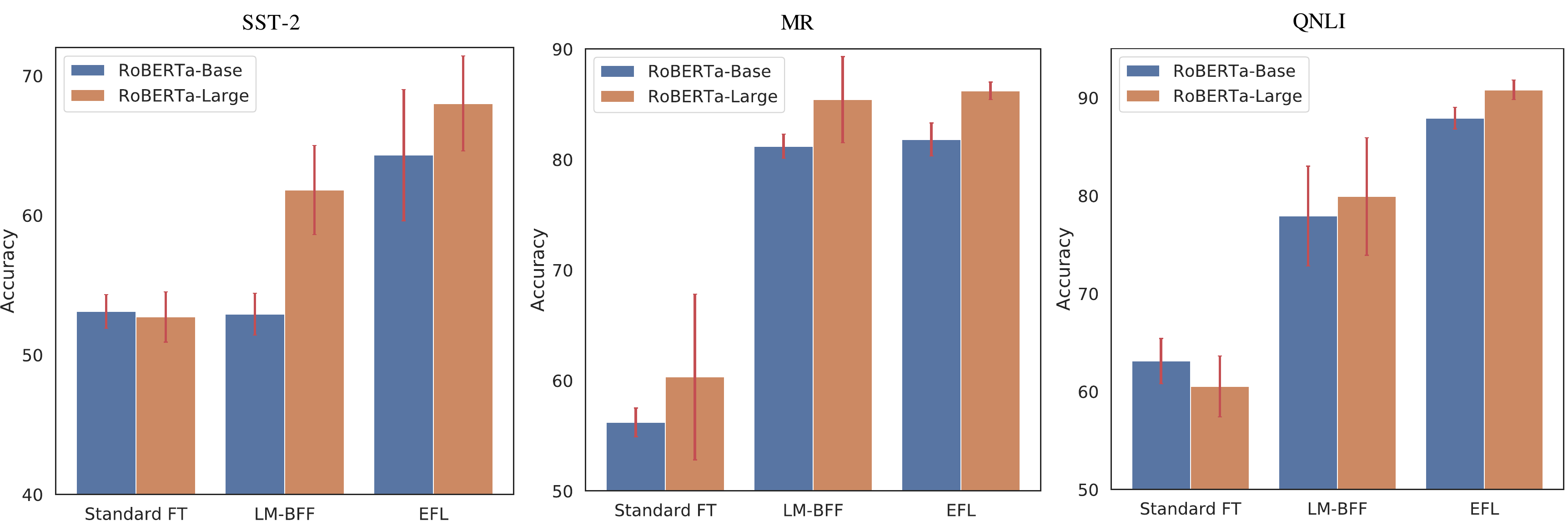}
  \caption{Impact of pre-trained Language model size on standard fine-tuning, LM-BFF and our proposed method EFL.}
  \label{fig:model_size}
\end{figure*}

\subsection{Impact of Training Data Scale}

We first investigate how our proposed method and other methods such as standard fine-tuning and prompt-based method LM-BFF scale as number of training data $K$ increases. In Figure~\ref{fig:sample_efficiency}, we plot the trends for SST-2 and QNLI. We observe that EFL has bigger improvements when number of annotated samples are small for both benchmarks. For simple tasks such as SST-2, when $K$=8, EFL already performs as good as $K$=256. For the harder tasks such as QNLI, EFL keep improving as $K$ increases, and perform best among all the methods. Another interesting observation is that, although EFL without entailment training performs poorly when $K=8$, it starts performing better than LM-BFF and standard fine-tuning when $K$ is larger than 16. Based on this observation, we further run ablation studies of our proposed methods without entailment training on two benchmarks: Subj and OS. In Table~\ref{table:sampel_efficiency_table}, we observe that the gap between EFL and EFL without entailment training is further reduced as $K$ increases and performs similarly when $K=32$.

\begin{table*}[h]
\centering
\renewcommand{\arraystretch}{1.1}
\resizebox{0.8\textwidth}{!}{
\begin{tabular}{lcccccc}
\toprule
 \textbf{Data} & \multicolumn{3}{c}{Few-shot $K$=16} & \multicolumn{3}{c}{Few-shot $K$=32}\\
& Standard FT & EFL wo PT & EFL & Standard FT & EFL wo PT & EFL\\
\midrule
Subj & 90.0 (1.7) & 90.4 (1.5) & \textbf{91.4} (1.3) & 92.0 (0.5) & \textbf{93.4} (0.9) & 92.6(0.9)\\
OS & 81.2 (2.1) & 84.1 (2.6) & \textbf{86.7} (2.0) & 88.0 (0.5) & \textbf{90.1} (0.9) & 88.8 (1.6)\\
\bottomrule
\end{tabular}}
\caption{Ablation studies of proposed EFL method without entailment training.}
\label{table:sampel_efficiency_table}
\end{table*}
\subsection{Impact of Model Size}

We further investigate the impact of model size on different methods. We evaluate standard fine-tuning, LM-BFF and our proposed method on two different-sized pre-trained LMs: RoBERTa-base (12 layer, 768 hidden dimension, 125M parameters) and RoBERTa-Large (24 layer, 1024 hidden dimension, 355M parameters). In Figure~\ref{fig:model_size}, we experiment with three benchmarks: SST-2, MR and QNLI. The performances of both LM-BFF and our proposed method improve when the size of pre-trained language models increases. Another interesting observation is that standard fine-tuning has better performance on SST-2 and QNLI dataset when uses a smaller language models. We suspect that this phenomenon is due to smaller LM has less number of parameters to be updated in standard fine-tuning method.

\section{Optimizations}

In this section, we discuss two further optimizations of our proposed framework: (i) a natural combination with unsupervised contrastive learning-based data augmentation method; (ii) multilingual few-shot learning.

\subsection{Unsupervised Contrastive Learning}

In our proposed framework, we reformulate various NLP tasks as a textual entailment task, which takes sentence pair as input. This approach facilitates leveraging unsupervised techniques to construct pairwise data to augment existing limited annotated data. Based on this observation, we propose a new data augmentation strategies, UCA, \textbf{U}nsupervised \textbf{C}ontrastive data \textbf{A}ugmentation. 

\begin{figure*}[t]
  \centering
  \includegraphics[width=\linewidth]{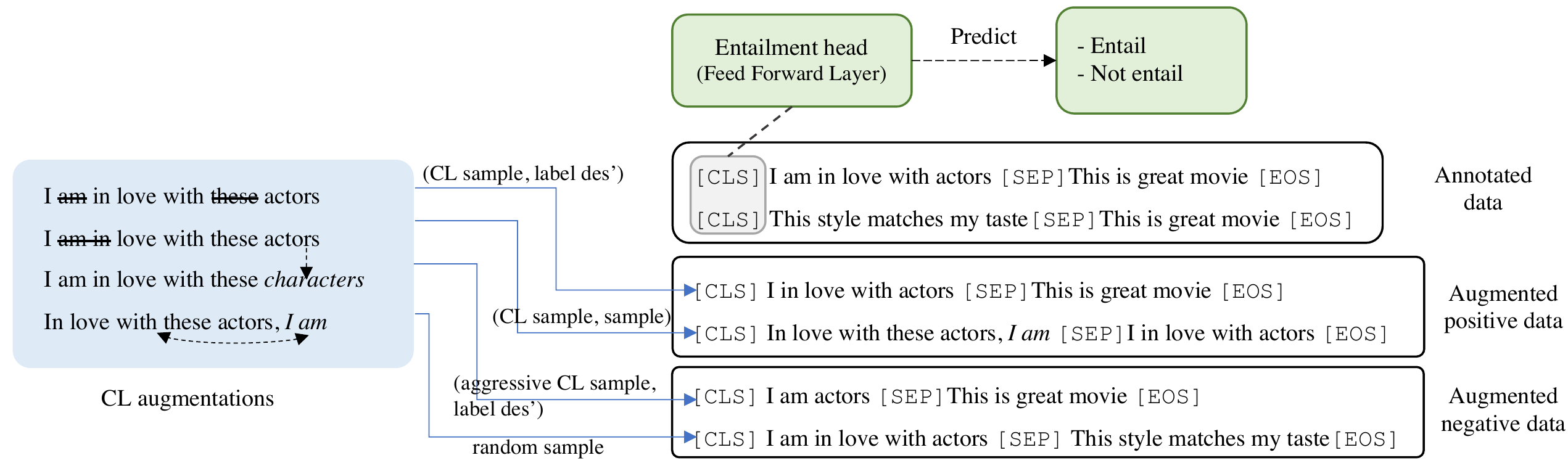}
  \caption{An illustration of our proposed unsupervised contrastive learning-based data augmentation method for entailment-based training.}
  \label{fig:uca}
\end{figure*}

The Figure~\ref{fig:uca} illustrates our main procedure of UCA. In EFL, each training sample has the format of $S_1 \texttt{[SEP]} S_2$, where $S_1$ is the original input sentence, $S_2$ is either label descriptions for one-sentence task or 2nd input sentence for sentence-pair task. In one-sentence task, we can construct a new instance $S_1'$ from $S_1$ via sentence augmentation, and add $S_1 \texttt{[SEP]} S_1'$ or $S_1' \texttt{[SEP]} S_2$ as a new training sample with positive label. Similarly, in sentence-pair task, we can also create a new positive instance $S_1 \texttt{[SEP]} S_1'$ or $S_2 \texttt{[SEP]} S_2'$. In practical implementation, we can alternate between these two augmentations in a probabilistic way. The negative sampling in UCA is straightforward: (i) we randomly sample two sentences $R_1$, $R_2$ from different instances of $D_{\text{train}}$ and construct a negative data $R_1 \texttt{[SEP]} R_2$; (ii) we use multiple sentence augmentation methods to largely change the original sentence meaning and add it as a negative sample. The rest relies on how to construct positive sentence pair from existing data.

We have four basic sentence augmentations: \texttt{word deletion}, \texttt{span deletion}, \texttt{reordering} and \texttt{substitution}. The basic intuition of deletion is that small portion of deletion in a sentence wouldn’t affect too much of the original semantic meaning. In word deletion, we randomly remove  $p_{\texttt{del}}$ percent of words in the input sentence, while in span deletion, we randomly pick $d_{\texttt{span}}$ consecutive words and directly delete them. Sentence reordering is originally proposed in BART~\citep{lewis2020bart} for auto-denoising, i.e., restoring original sentence from random reordered sentence. We randomly sample $d_{\texttt{re}}$ pairs of span and switch them pairwise to construct the reordering augmentation in our implementation. In synonym substitution~\citep{jia2019certified}, we sample $d_{\texttt{sub}}$ words and replace them with synonyms to construct one augmentation. This approach is known to be able to improve model’s robustness, which is beneficiary for some grammatical acceptability task such as CoLA~\citep{warstadt2019neural}.

\begin{table*}[h]
\renewcommand{\arraystretch}{1.1}
\resizebox{\textwidth}{!}{
\begin{tabular}{lcccccccc}
\toprule
 \textbf{Method} & \textbf{SST-2} & \textbf{MR} & \textbf{CR} & \textbf{MPQA} & \textbf{Subj} & \textbf{IMDB} & \textbf{OS}&  \textbf{CoLA}\\ 
& (Acc) & (Acc) & (Acc) & (Acc) & (Acc) & (Acc) & (Acc) & (Acc.) \\
\midrule
 EFL & \textbf{90.8} (1.0) & 86.2 (0.8) & 92.3 (0.4) & 87.0 (0.6) & 80.0 (5.4) & 87.1 (1.2) & 76.6 (3.5) & 69.4 (0.9)\\
 EFL + UCA & 90.3 (1.2) & \textbf{86.3} (1.0) & \textbf{92.8} (0.3) & \textbf{87.8} (0.5) & \textbf{83.3} (1.2) & 87.4 (1.2) &  \textbf{79.8} (3.3) & \textbf{71.2} (2.1)\\\midrule
 \textbf{Method} & \textbf{QQP} & \textbf{QNLI} & \textbf{SNLI} & \textbf{RTE} & \textbf{MRPC} & \textbf{STS-B} & \textbf{BoolQ} & \textbf{Average} \\
 & (F1) & (Acc) & (Acc) & (Acc) & (F1) & (Pear.) & (Acc) \\
\midrule
EFL & 67.3 (2.6)  & 68.0 (3.4) & \textbf{81.0} (1.1) & 85.8 (0.9) & 76.2 (1.3) & 71.0 (1.3)  & 73.9 (1.8) & 79.5 \\
  EFL + UCA & \textbf{81.5} (0.8)  & \textbf{74.6} (0.8) & 80.9 (1.1) & \textbf{87.2} (0.9)  & \textbf{80.8} (0.8) & \textbf{75.7} (1.7) & 73.7 (1.9) & \textbf{82.2}\\
\bottomrule
\end{tabular}}
\caption{Few-shot learning results ($K=8$) with EFL and unsupervised contrastive data augmentation. Note that bold numbers refer to the new best results achieved compared to results in Table~\ref{table:few_shot_16_1} and Table~\ref{table:few_shot_16_2}.}
\label{table:uca}
\end{table*}

\begin{table*}[h]
\renewcommand{\arraystretch}{1.1}
\resizebox{\textwidth}{!}{
\begin{tabular}{lccccccccc}
\toprule
\textbf{Method} & \textbf{MPQA} & \textbf{Subj} & \textbf{IMDB} & \textbf{OS} & \textbf{CoLA} & \textbf{QQP} & \textbf{QNLI}&  \textbf{STS-B} & \textbf{Average}\\ + UCA 
& ($\Delta$Acc) & ($\Delta$Acc) & ($\Delta$Acc) & ($\Delta$Acc) & ($\Delta$Acc) & ($\Delta$F1) & ($\Delta$Acc) & ($\Delta$Pear.) & - \\
\midrule
Standard FT & -2.05  & +4.46  & -2.26  & -2.58  & +1.60  & +17.5  & +11.9 & +6.21 & +4.35\\
LM-BFF & +0.10 & +\textbf{13.9} & -0.76 & -1.90 & +2.89 & +7.79 & +2.62 &  +2.35 & +3.38\\
EFL & +0.81  & +3.29  & +0.28  & +3.17  & +1.85  & +14.2  & +6.57 & +5.57 & +4.47 \\
\bottomrule
\end{tabular}}
\caption{Results of combining unsupervised contrastive data augmentation (UCA) with different methods. We report average $\Delta$ across 5 different training sets. Note that bold numbers refer to the new best results achieved compared to results in Table~\ref{table:few_shot_16_1} and Table~\ref{table:few_shot_16_2}.}
\label{table:ablation_cl}
\end{table*}
As illustrated in Table~\ref{table:uca}, UCA further brings significant improvement across 12 of 15 tasks, and average \textbf{2.7pt} improvements across 15 tasks. In particular, the improvement is even larger in the sentence-pair task. For example, we have seen 14pt improvement of QQP task. This is intuitive since UCA creates more similarly distributed data via sentence-level augmentation. In the linguistic acceptability task CoLA, the UCA can create some negative data by aggressively delete or reorder words of sentences, which matches the data format of this task.

It is also interesting to see how our proposed UCA method impacts other few-shot or fine-tuning methodologies. In Table~\ref{table:ablation_cl}, we further make a direct comparison of combining UCA with standard fine-tuning, LM-BFF and our proposed method on 5 single-sentence task and 3 sentence-pair task. As it is shown, the UCA method improves all 8 tasks for EFL, 5 of 8 task for standard fine-tuning, 6 of 8 tasks for LM-BFF, and brings average 3 to 4pt improvement. It implies our proposed method can benefit different training methods and may not be limited to EFL. Further, we observe that combining UCA method with standard fine-tuning and LM-BFF mostly improves sentence-pair task. The major reason is that only sentence-pair task in standard fine-tuning and LM-BFF has similar input format of UCA data, instead, the training sample created by EFL is similar format as UCA data.

\subsection{Multilingual Few-shot Learning}

When we scale an existing problem to a new language, one challenge is how to obtain the annotated data in the target domain, which leads to an even important research area, \textit{multilingual few-shot learning} (MFL). In MFL, suppose that we have language space $\mathcal{L}=\{l_1, \ldots, l_m\}$ with $m$ different languages. The training dataset is similarly defined as $D_{\text{train}}$ as in previous monolingual scenario. The key difference is the test dataset,
\begin{equation*}
D_{\text{test}}^X=\{D_{\text{test}}^{l_1},D_{\text{test}}^{l_2},\ldots,D_{\text{test}}^{l_m}\},
\end{equation*}
where $D_{\text{test}}^{l_m}$ is the test dataset in language $l_m$. This problem is even challenging compared to previous scenario since we not only need to generalize to the original unseen test data, but also generalize to unseen languages.

\begin{figure*}[h]
  \centering
  \includegraphics[width=\linewidth]{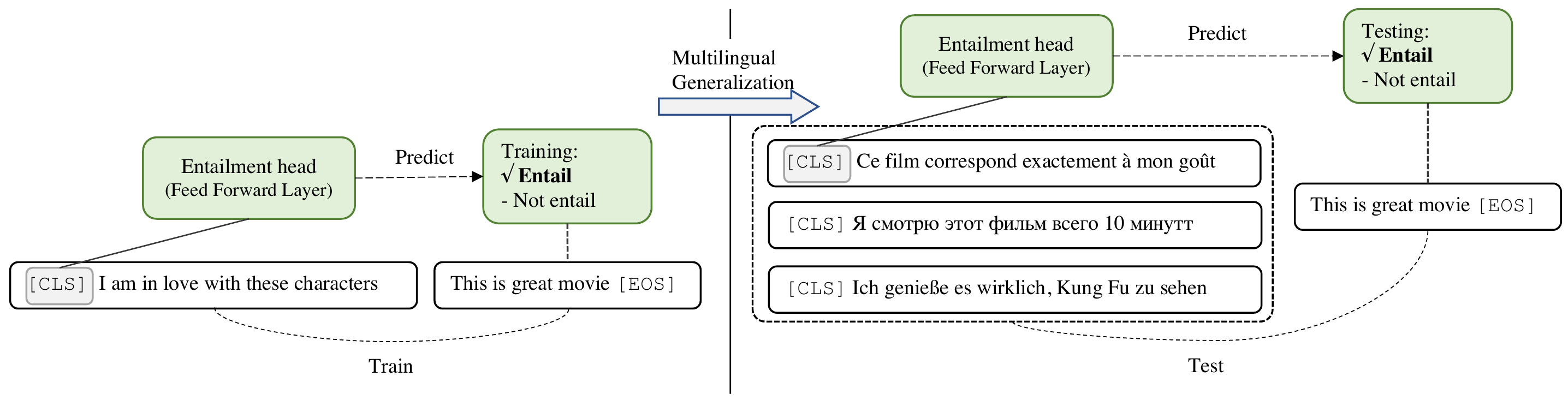}
  \caption{Generalization of our proposed method to multilingual few-shot learning: we first fine-tune on English data with few annotated examples and then test it on multilingual data.}
  \label{fig:multilingual}
\end{figure*}

There have been existing works leveraging multilingual masked language modeling to develop cross-lingual LM such as XLM~\citep{lample2019cross}. However, there are no existing works on exploring MFL. As shown in Fig~\ref{fig:multilingual}, our proposed method can be straightforwardly extended to this scenario. We follow the same steps when we construct the training dataset in monolingual scenario. In test dataset $D_{\text{test}}^X$, we don't need to translate label description into target language, instead, simply use original English label description for various languages.

We follow the same evaluation protocol in the monolingual scenario. For evaluation benchmark, we reuse the previous monolingual benchmark by using the translation test~\citep{ott2018scaling}. We also use XLM-R~\citep{conneau2020unsupervised} as our pre-trained LM. For our proposed method EFL, we fine-tune the model on XNLI dataset~\citep{conneau2018xnli} instead of MNLI. Table~\ref{table:multilingual_few_shot} shows the main results of 7 NLP tasks. We observe that EFL performs significantly better in the multilingual scenarios compared to standard fine-tuning, i.e, improved average accuracy from 61.9 to 80.7 across 7 tasks. This result demonstrates the effectiveness of EFL on multilingual scenario. We further experiment with translate training method: translate few-shot English annotated samples into target languages and add them into training data. Interestingly, we observe standard fine-tuning method improves from 61.9 to 67.0, while EFL only improves from 80.7 to 81.3.

\begin{table*}[t]
\renewcommand{\arraystretch}{1.1}
\resizebox{\textwidth}{!}{
\begin{tabular}{lcccccccc}
\toprule
 \textbf{Method} & \textbf{M-SST-2} & \textbf{M-MR} & \textbf{M-CR} & \textbf{M-MPQA} & \textbf{M-Subj}  & \textbf{M-OS}&  \textbf{M-CoLA} &\textbf{Average}\\ 
  & (Acc) & (Acc) & (Acc) & (Acc) & (Acc) & (Acc) & (Acc) & (Acc) \\
\midrule
Majority & 50.9 & 50.0 & 50.0 & 50.0 & 50.0  & 66.8 & 69.1 & 55.3\\
\midrule
\multicolumn{6}{l}{\textit{Train + Translate Test + Few-shot $K=8$}}\\
\midrule
Standard FT & 52.2 (0.8) & 54.3 (1.4) & 55.8 (1.0) & 56.9 (2.1) & 77.5 (4.7) & 67.2 (0.8) & 69.2 (0.1) & 61.9\\
 EFL wo PT & 52.6 (1.6) & 54.3 (1.5) & 57.2 (3.4) & 59.9 (1.8) & 76.5 (6.4) & 67.0 (0.4) & 69.2 (0.1) & 62.4\\
 EFL & 84.1 (1.1) & 81.2 (0.8) & 90.1 (0.4) & 83.2 (0.5) & 84.5 (2.0) & 72.9 (2.7) & 68.7 (1.2) & 80.7\\
 \midrule
\multicolumn{6}{l}{\textit{Translate Train + Translate Test + Few-shot $K=8$}}\\
\midrule
Standard FT & 55.1 (5.0) & 63.8 (7.2) & 65.5 (4.9) & 57.0 (1.6) & 89.4 (1.6) & 69.4 (7.4) & 69.1 (0.1) & 67.0\\
 EFL wo PT & 54.6 (3.2) & 60.7 (4.9) & 65.6 (4.1) & 59.3 (3.3) & \textbf{89.5} (1.3) & 71.3 (5.4) & \textbf{69.2} (0.1) & 67.2\\
 EFL & \textbf{84.3} (0.8) & \textbf{81.5} (0.9) & \textbf{90.5} (1.1) & \textbf{83.3} (1.3) &  88.7 (2.1) & \textbf{73.1} (2.1) & 67.6 (1.6) & \textbf{81.3}\\
\bottomrule
\end{tabular}}
\caption{Our main multilingual few-shot learning results using XLM-R. All the results are evaluated on full test sets and averaged across 5 different training sets. We add M- in front of each benchmark to represent the multilingual testing.}
\label{table:multilingual_few_shot}
\end{table*}

\section{Conclusion}

In this paper, we proposed a set of simple and effective few-shot learning methods: (i) reformulates traditional classification/regression tasks as a textual entailment task; (ii) unsupervised contrastive learning-based data augmentations. We show through a series of systematic evaluations that our method outperforms various few-shot learning methods by up to 55\% (and 12\% on average). In the future, we will explore several new directions based on this approach: (i) how to choose the best label descriptions based on reinforcement learning; (ii) how to create a more effective entailment training task instead of MNLI tasks.

\bibliography{reference}
\bibliographystyle{iclr2020}

\appendix
\section{Experiment Details\label{appendix:experiment}}

The label descriptions we use for each method are the same and defined in Table~\ref{table:label_prompt}. For model hyperparameters, in full data experiment, we use the learning rate 1e-5, batch size 32, weight decay 0.1, max epochs 10 and linear learning rate decay with a warm up ratio 0.06. In few-shot experiments and ablation studies, we use a constant  learning rate 1e-5, batch size 8, max epochs 10, and standard Adam optimizer.

For unsupervised contrastive data augmentation (UCA), we use the following default setups: we randomly augment 8 data for each class. 
\begin{itemize}
    \item Positive generation: We alternate sentence augmentations with 10\% probability in deleting character, 10\% probability in reordering words, 40\% probability in deleting words, and 40\% probability in reordering words. In deleting characters, we delete 15\% characters of input sentence. In reordering spans, each span cross 5\% consecutive characters of input sentence, and we consider switching of 3 pairs of spans. In deleting words, we randomly delete 15\% words of input sentence. In reordering words, we randomly switch 2 pairs of words.
    \item Negative generation: It has same probability distribution across different sentence augmentations but higher probability of reordering and deletion: In deleting characters, we delete 40\% characters of input sentence; In reordering spans, each span cross 25\% consecutive characters of input sentence, and we consider switching of 3 pairs of spans. In deleting words, we randomly delete 40\% words of input sentence. In reordering words, we randomly switch 2 pairs of words and each contains 40\% words.
\end{itemize}
For one-sentence task, suppose $p_1$ is the label description, we use the above rule to generate positive sample $S_1'$ from input sentence $S_1$ and add augmented data: $S_1\texttt{[SEP]}S_1'$ or $S_1'\texttt{[SEP]}p_1$ with equal probability. The negative augmented data consists of 70\% of randomly pairwise down-sampled data and and 30\% UCA constructed negative samples. In pairwise down-sampling, we randomly sample the first sentence $S_{i1}$ from existing positive sample, and then randomly sample another first sentence $S_{j1}$ from existing negative sample, and construct $S_{i1}\texttt{[SEP]}S_{j1}$ as new sample. For sentence-pair task, suppose that we have one annotated sample $S_1\texttt{[SEP]}S_2$, we will add augmented data $S_1\texttt{[SEP]}S_2'$ or $S_1'\texttt{[SEP]}S_2$ with equal probability. The negative data follows the similar rule of one-sentence task.

\begin{table*}[htb]
\renewcommand{\arraystretch}{1.1}
\resizebox{\textwidth}{!}{
\begin{tabular}{cll}
\toprule
\textbf{Dataset}& \textbf{Template of Prompt Finetuning} & \textbf{Template of EFL} \\\midrule
SST-2 & sentence$_1$\texttt{[SEP]}It was\texttt{[MASK]}. (great / terrible) & sentence$_1$\texttt{[SEP]}It was great\\
MR & sentence$_1$\texttt{[SEP]}It was\texttt{[MASK]}. (great / terrible) & sentence$_1$\texttt{[SEP]}It was great\\
CR & sentence$_1$\texttt{[SEP]}It was\texttt{[MASK]}. (great / terrible) & sentence$_1$\texttt{[SEP]}It was great\\
MPQA & sentence$_1$\texttt{[SEP]}It was\texttt{[MASK]}. (positive / negative) & sentence$_1$\texttt{[SEP]}It was positive\\
Subj & sentence$_1$\texttt{[SEP]}It was\texttt{[MASK]}. (subjective / objective) & sentence$_1$\texttt{[SEP]}It was objective\\
OS & sentence$_1$\texttt{[SEP]}It was\texttt{[MASK]}. (hatespeech / benign) & sentence$_1$\texttt{[SEP]}It was hatespeech\\
IMDB & sentence$_1$\texttt{[SEP]}It was\texttt{[MASK]}. (great / terrible) & sentence$_1$\texttt{[SEP]}It was great\\
CoLA & sentence$_1$\texttt{[SEP]}It was\texttt{[MASK]}. (correct / incorrect) & sentence$_1$\texttt{[SEP]}It was correct\\
\multirow{2}{*}{Yelp} & \multirow{2}{*}{ \makecell[l]{sentence$_1$\texttt{[SEP]}It was\texttt{[MASK]} news.\\(Great/good/ok/bad/terrible)}}  & \multirow{2}{*}{ \makecell[l]{sentence$_1$\texttt{[SEP]}It was\\(great/good/ok/bad/terrible).}} \\
& & \\
\multirow{2}{*}{AG news} & \multirow{2}{*}{ \makecell[l]{sentence$_1$\texttt{[SEP]}It was\texttt{[MASK]} news.\\(World/sports/business/science)}}  & \multirow{2}{*}{ \makecell[l]{sentence$_1$\texttt{[SEP]}It is (World/sports/\\business/science) news.}} \\
& & \\
\multirow{2}{*}{Trec} & \multirow{2}{*}{ \makecell[l]{sentence$_1$\texttt{[SEP]}It was\texttt{[MASK]} news.\\(expression/entity/description/human/location/number)}}  & \multirow{2}{*}{ \makecell[l]{sentence$_1$\texttt{[SEP]}It is (expression/\\entity/description/human/location/number).}} \\
& & \\
\midrule
QQP & sentence$_1$ ?\texttt{[MASK]}, sentence$_2$. (yes / no) & sentence$_1$\texttt{[SEP]}sentence$_2$\\
MRPC & sentence$_1$ ?\texttt{[MASK]}, sentence$_2$. (yes / no) & sentence$_1$\texttt{[SEP]}sentence$_2$\\
QNLI & sentence$_1$ ?\texttt{[MASK]}, sentence$_2$. (yes / no) & sentence$_1$\texttt{[SEP]}sentence$_2$\\
MNLI & sentence$_1$ ?\texttt{[MASK]}, sentence$_2$. (yes / maybe / no) & sentence$_1$\texttt{[SEP]}sentence$_2$\\
SNLI & sentence$_1$ ?\texttt{[MASK]}, sentence$_2$. (yes / maybe / no) & sentence$_1$\texttt{[SEP]}sentence$_2$\\
RTW & sentence$_1$ ?\texttt{[MASK]}, sentence$_2$. (yes / no) & sentence$_1$\texttt{[SEP]}sentence$_2$\\
STS-B & sentence$_1$ ?\texttt{[MASK]}, sentence$_2$. (yes / no) & sentence$_1$\texttt{[SEP]}sentence$_2$\\
BoolQ & sentence$_1$ ?\texttt{[MASK]}, sentence$_2$. (yes / no) & sentence$_1$\texttt{[SEP]}sentence$_2$\\
\bottomrule
\end{tabular}}
\caption{Prompts and label descriptions of prompt-based finetuning method~\citep{gao2020making} and our method used in experiments. }
\label{table:label_prompt}
\end{table*}
\section{Benchmark}
Our benchmark includes 8 datasets from GLUE~\citep{wang2018glue}: CoLA~\citep{warstadt2019neural}, SST-2~\citep{socher2013recursive}, MPRC~\citep{dolan2005automatically}, QQP\footnote{https://www.quora.com/share/First-Quora-Dataset-Release-Question-Pairs}, STS-B~\citep{cer2017semeval}, MNLI~\citep{williams2017broad}, QNLI~\citep{rajpurkar2016squad}, RTE~\citep{dagan2005pascal,haim2006second,giampiccolo2007third, bentivogli2009fifth}, IMDB~\citep{maas2011learning}, Yelp, AG News~\citep{zhang2015character}, SNLI~\citep{bowman2015large}. For the datasets which require a cross-validation evaluation, we follow similar processing path~\citep{gao2020making}: MR~\citep{pang2005seeing}, CR~\citep{hu2004mining}, MPQA~\citep{wiebe2004annotating}, Subj~\citep{pang2004sentimental}—we simply randomly sample 2,000 examples as the testing set and leave them out from training. BoolQ~\citep{clark2019boolq} is from SuperGlue~\citep{wang2019superglue}, and OS is twitter offensive speech data~\citep{davidson2017automated}.

\end{document}